\documentclass[letterpaper, 10 pt, conference]{ieeeconf}  %

\IEEEoverridecommandlockouts                              %
\usepackage{graphicx}
\usepackage{amsmath}
\usepackage{amssymb}
\usepackage{algorithm}
\usepackage{algpseudocode}
\usepackage{capt-of}
\usepackage{xcolor}
\makeatletter\let\NAT@parse\undefined\makeatother   %
\usepackage[numbers,sort&compress]{natbib}   %
\usepackage{xr-hyper}                        %
\usepackage[hidelinks]{hyperref}             %

\title{\LARGE \bf
PHIRL: Aligning Learned Rewards with Task Progress for Inverse Reinforcement Learning
}

\author{
Hang Yu$^{*}$, James Staley, Cheng Xi Tsou, Xiujin Liu, Wenchang Gao, Jindan Huang, \\
Shijie Fang, Zhegong Shangguan, Angelo Cangelosi, Reuben M Aronson, Elaine Short
\thanks{$^{*}$Corresponding author: {\tt\small hyu08@tufts.edu}}
}

\begin{document}
\maketitle
\thispagestyle{empty}
\pagestyle{empty}

\begin{abstract}
Human demonstrations provide dense policy-level information but sometimes lack local precision. Human feedback presents accurate local critiques, but offers sparse evaluations rather than direct policy guidance. 
We propose \textit{Progress}-Heuristicized Inverse Reinforcement Learning (PHIRL), a data-efficient framework that learns robust reward functions by jointly leveraging demonstrations and feedback. 
Specifically, we use \textit{progress}, a feedback modality that describes cumulative task completion.
PHIRL iteratively infers a reward function from demonstrations via inverse reinforcement learning, calculates the learned rewards over the \textit{progress}-annotated demonstrations, and aligns the rewards with \textit{progress} annotations over four dimensions. 
We evaluate PHIRL on real and simulated robot tasks, with additional exploration using a fine-tuned vision-language model to provide \textit{progress} feedback. 
Results demonstrate that PHIRL significantly outperforms the baselines, achieving substantially higher environmental return rewards and task success with only twenty percent of demonstrations annotated. 
Analysis of reward-hacking scenarios demonstrates that PHIRL’s learned reward functions are reliable against exploitation.

\end{abstract}

\section{Introduction}
An appropriate reward function is a crucial component for Reinforcement Learning (RL) agents to perform effectively \cite{ibrahim2024comprehensiveoverviewrewardengineering, 11127866}.
While sparse rewards are intuitive to specify, they offer limited guidance during early exploration. Conversely, hand-crafted dense rewards accelerate learning, but designing them requires substantial engineering effort and tedious manual tuning \cite{ibrahim2024comprehensiveoverviewrewardengineering}. 
Inverse Reinforcement Learning (IRL) offers an alternative by inferring a dense reward function directly from human demonstrations or human feedback, mitigating the need for manual reward designing.
Compared to imitation learning methods that directly map states to actions \cite{chi2023diffusion}, IRL may require higher-quality demonstrations \cite{sasaki2020behavioral}, but the learning outcomes can be easier for post-training \cite{sun2025inversereinforcementlearningmeets} or revising \cite{hao2026rewardtransferinversereinforcement}.
In this work, we aim to learn robust reward functions by combining human demonstrations and feedback, while reducing human engagement.

Recent work has demonstrated that using human demonstrations and feedback jointly can improve data efficiency \cite{li2024learningrewardpolicyjointly}
 and lower quality requirements for human demonstrations \cite{palan2019learning}. 
Many reward learning methods often assume the demonstrations are optimal \cite{ ayalew2025progressor} or near-optimal \cite{sasaki2020behavioral}, but providing high-quality demonstrations for manipulation tasks is known to be difficult even for experts. 
High-quality human feedback, on the other hand, is more accessible and requires fewer skills. 
Prior reward learning work often relies on preference feedback \cite{palan2019learning, ibarz2018reward, brown2019ranking, gao2026steplevelpreferencelearninggenerative}.
Forcing users to pick a preferred trajectory does not always benefit learning. For instance, consider a pick-and-place task where the robot never manages to grasp the object. In this case, all sampled trajectory clips are undesirable. The preference feedback is therefore uninformative.

Moreover, prior work has achieved huge success in simulating human feedback \cite{chen2026topreward, yu2026chi0resourceawarerobustmanipulation, 11217604}, but for scenarios where feedback from real human users is required \cite{zhou2026mindsim2realgapuser}, users often need to stay in the training loop to provide feedback on newly generated trajectories. 
Technologies and dataset-sharing communities, such as UMI \cite{chi2024universalmanipulationinterfaceinthewild}, ego-centric data \cite{kareer2025egomimic}, and Hugging Face \cite{jain2022hugging}, have made robot demonstrations significantly easier to access.
In this setting, human-in-the-loop teaching effort has become a more expensive resource than stock demonstrations.

In this work, we mitigate these gaps by proposing a data-efficient reward learning framework, PHIRL, that learns from demonstrations and human \textit{progress} \cite{yu2025progressdidimake}.
\textit{Progress} describes the cumulative task completion based on the task observation. 
Our key intuition is that \textbf{the rewards inferred from demonstrations should be consistent with the annotations of the demonstrations.} 
This process mimics human learning: much like a student cross-referencing answers to correct their own solutions, the method generates a potential reward function and compares it against human feedback to refine its learned rewards.
In PHIRL, we iteratively use an IRL algorithm to learn a reward function, calculate rewards over the annotated demonstrations, and then shape the learned reward function in four dimensions: temporal delta, magnitudes, potential, and completion. In addition, we extend PHIRL by integrating a fine-tuned Vision-Language Model (VLM) to provide online \textit{progress} feedback.

We validate PHIRL with two robot tasks from Robosuite \cite{zhu2025robosuitemodularsimulationframework} and one real robot task with four baselines.  
Our results show that PHIRL significantly outperforms IRL baselines 
with only 20\% of demonstrations being annotated offline, and the online extension of PHIRL is effective with VLM-generated annotations. 
We conducted a qualitative analysis against AIRL, which confirms that the reward functions shaped by PHIRL are resilient to two reward-hacking behaviors.
Code is available at github.com/PHIRL2026.

\begin{figure*}
    \centering
    \includegraphics[width=.95\linewidth]{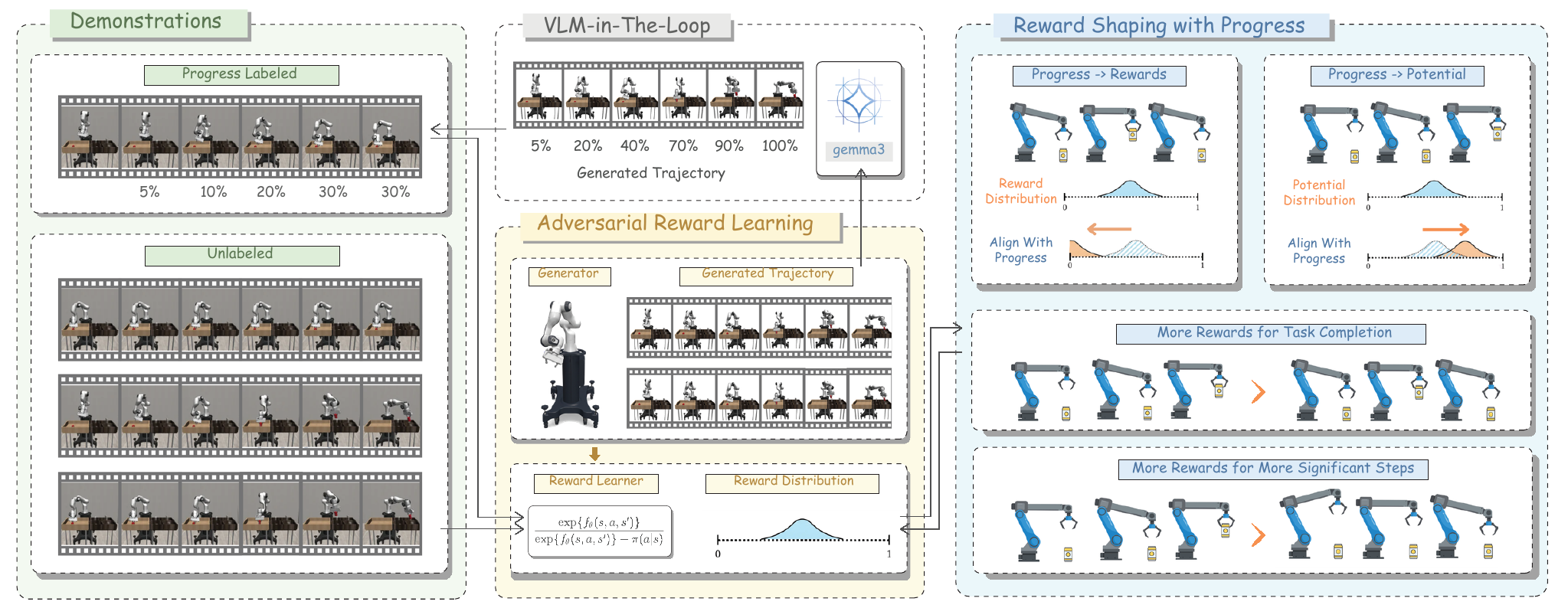} 
    \caption{PHIRL framework.  We randomly sample a subset of demonstrations and ask users to annotate them with \textit{progress}. We then iteratively use an IRL algorithm to learn a reward model from the demonstrations and align the learned reward model with \textit{progress} over four dimensions. 
    Optionally, the generated trajectories will be sampled and annotated by a VLM for data augmentation. 
    The generated trajectories will be used for reward shaping only. 
    }
    \label{fig:pip}
\end{figure*}

\section{Background}
Interactive machine learning grants robots the ability to adapt to human needs or learn new skills by leveraging human knowledge \cite{arzate2020survey}, such as human preference \cite{akrour2011preference}, facial expressions \cite{cui2021empathic}, numerical evaluation \cite{knox2009interactively}, and human demonstrations \cite{mehta2024unified}. 

\noindent \textbf{Reward Learning from Human Feedback}
Human feedback is a popular way for agents to learn from humans via interactive reinforcement learning or inverse reinforcement learning \cite{mehta2024unified, cui2021empathic, yu2023thumbs}. 
Human feedback contains relatively sparse but precise information \cite{palan2019learning}.
To provide feedback,
humans first observe agents performing a task, and then evaluate the performance of the agents implicitly or explicitly \cite{arzate2020survey}.
Humans can use feedback to assess robot behaviors \cite{gunjal2025rubrics, yu2021active} and indicate their preferred policy, which can then be used to extract rewards \cite{ zhang2025rewind}. 
Human feedback can also be applied to post-train trained models, such as foundation models \cite{sun2025inversereinforcementlearningmeets}.

\noindent \textbf{Reward Learning from Human Demonstrations}
Learning from Demonstration (LfD) allows learning agents to learn to perform new tasks by imitating humans \cite{ mehta2024unified,    sasaki2020behavioral, 11217746}.
LfD approaches have many advantages, such as reducing the need for expert programming \cite{zhu2018robot} and data efficiency \cite{chi2023diffusion}.
LfD generally solves the learning problem mainly in two ways: (1) inferring a policy \cite{ chi2023diffusion, hu2024thought}, and (2) inferring a reward function \cite{ren2024hybrid,  fu2018learning}.
Learning a reward function from demonstrations, while more difficult, has multiple advantages: it enables the learning outcome to be transferred to new tasks \cite{Yoo_2022}, supports rapid re-optimization \cite{kim2024unifiedlinearprogrammingframework}, and can be combined with optimal control and RL machinery \cite{zhang2024improvingreinforcementlearninghuman}.  
However, learning an accurate reward function purely from demonstrations is challenging \cite{palan2019learning}, and the majority of demonstrations often need to be high-quality \cite{sasaki2020behavioral, duan2026learning}.

\noindent \textbf{Reward Learning from Both Human Demonstration and Human Feedback}
Recent work indicates that combining human demonstrations and human feedback could improve learning efficiency and overcome the limitations of using human feedback or demonstrations alone \cite{ibarz2018reward, brown2020safe, palan2019learning, mehta2024unified}. 
Specifically, \citet{ibarz2018reward} proposed to use demonstrations to pre-train a learning agent, and then used the pre-trained agent to generate preference queries, which significantly improves the sample efficiency for preference learning. 
Building on \citet{ibarz2018reward}'s work, \citet{palan2019learning} not only used the demonstrations to improve sample efficiency but also used the demonstrations to learn an initial reward model. 
Two related works \cite{ayalew2025progressor, yang2024rank2reward}  showed that learned reward functions can be improved by forcing the models to assign more rewards to expert demonstrations. However, this improvement is based on the assumption that expert demonstrations always make at least some task progress.
\citet{kim2024unifiedlinearprogrammingframework, li2024learningrewardpolicyjointly} showed that the learned reward function can be improved by aligning with human preference comparisons.

Our work differs from prior work by improving the data efficiency and reducing human-in-the-loop effort. 
We do not assume that behaviors in demonstrations always contribute to task completion, and we align the learned rewards with \textit{progress} across multiple dimensions rather than temporal delta alone.   
We align the learned rewards with \textit{progress} annotations collected prior to learning, thereby making them more robust to non-progressive behaviors while requiring less human engagement.

\section{Methodology}
Our goal is to effectively learn \textit{robust} reward functions. 
As previously mentioned, human demonstrations can be combined with human feedback to improve learning efficiency.
Our work is built on the key insight that \textbf{rewards inferred from a set of demonstrations should be consistent with the feedback on the same set of demonstrations.}

\subsection{Progress}
We use a numerical human feedback form, \textit{progress} \cite{yu2025progressdidimake} for reward function alignment. 
\textit{Progress} is a signal that describes the cumulative degree of task completion, ranging from fully incomplete to fully complete.
Previous work has demonstrated that progress is an informative teaching signal and has many advantages for reward learning \cite{zhang2026progresslmprogressreasoningvisionlanguage, yu2026chi0resourceawarerobustmanipulation}.
For any given task $t$, current state $s$, any action $a_i$ and $a_j$, and any previous state $s_i$ and $s_j$, the \textit{progress}:
\begin{equation}
\label{eq:prog_state_only}
\mathrm{prog}_t(s)
\;=\;
\mathrm{prog}_t\!\bigl(s \mid s_i,a_i\bigr)
\;=\;
\mathrm{prog}_t\!\bigl(s \mid s_j,a_j\bigr),
\quad \forall i,j,
\end{equation}

In this work, we use \textit{progress} in a range of 0 to 100 and ask participants to provide \textit{progress} labels only based on the current observation, where 100 indicates the task is fully complete, and 0 means that no progress has been made yet. 
\textit{Progress} is accurate for low-quality demonstrations and consistent across non-experts without requiring additional cognitive load compared to providing preference evaluations \cite{yu2025progressdidimake}.

\subsection{Reward Learning from Demonstrations}
Given a set of human demonstrations $\mathcal{D} = \{ d_0, ..., d_n\}$, where $d = (s_0, a_0, ..., s_n, a_n)$,
the objective of IRL is to find a reward function $r(s, a, s')$.
If we assume that the demonstrations are from an optimal policy $\pi^*$, we can then interpret the IRL problem as a maximum likelihood problem:
\begin{equation}
\label{eq1}
    \max_{\theta} \mathbb{E}_{d \sim \mathcal{D}} [\log \mathbb{P}_{\theta}(d)],
\end{equation}
where 
\begin{equation*}
\begin{aligned}
\mathbb{P}_{\theta}(d) \propto{}& \mathbb{P}(s_0) \left( \prod_{t=0}^{T-1} \mathbb{P}(s_{t+1} \mid s_t, a_t) \right) \\
&\times \exp \left( \sum_{t=0}^{T-1} \gamma^t r_{\theta}(s_t, a_t) \right)
\end{aligned}
\end{equation*}
We use the Adversarial Inverse Reinforcement Learning (AIRL) \cite{fu2018learning} algorithm to learn the initial reward distribution.
AIRL casts the optimization of equation \ref{eq1} as a GAN \cite{goodfellow2020generative} optimization problem.
The discriminator uses a particular $f_\theta$:
\begin{equation}
D_{\theta}(s, a, s') = \frac{\exp\{f_{\theta}(s, a, s')\}}{\exp\{f_{\theta}(s, a, s')\} + \pi(a|s)}
\end{equation}
and $\pi$ is trained to maximize:$ r =  \log D_\theta - \log(1 - D_\theta)$,
where $f_{\theta}(s, a, s_0)$ can be interpreted as the advantage under deterministic dynamics:
\begin{equation}
    f^{*}(s, a, s') = \underbrace{r^{*}(s) + \gamma V^{*}(s')}_{Q(s,a)} - \underbrace{ V^{*}(s)}_{V(s)} = A^{*}(s, a)
\end{equation}
We select AIRL as our reward learning method over other methods because
AIRL infers reward functions from state-only functions  $r(s)$ and $V(s)$, and \textit{progress} is also path-independent.

\subsection{Reward Function Shaping}
We improve the learned reward function by shaping it to align with \textit{progress} annotations during reward learning. 
Specifically,
we exploit four key properties of the information provided by \textit{progress}.

\noindent \textbf{Temporal Delta}: positive changes in \textit{progress} correspond to positive rewards.
Given a trajectory segment $s_{t} \xrightarrow {a_t} s_{t+1}$, if the robot's action sequence is appropriate, the \textit{progress} $p_{t+1}$ at state $s_{t+1}$ should be higher than the \textit{progress} $p_{t}$ at $s_{t}$. Similarly, the learned incremental reward $r_\theta(s_{t}, s_{t+1}) = r_\theta(s_{t+1}) - r_\theta(s_t)$ should be positive, encouraging transitions that move the robot closer to task completion.  We enforce this relationship via the following loss:
\begin{equation}
\label{eq:pdp,pr}
\mathcal{L}_{\Delta p \rightarrow r_\theta}(\theta)
=\;
\mathbb{E}_{(s_t, s_{t+1})\sim\mathcal{D}}
\!\Bigl[
\,\ell\bigl(\Delta p_t,\,r_\theta(s_t,s_{t+1})\bigr)
\Bigr],
\end{equation}
where $\Delta p_t = p_{t+1} - p_t$ , and $\ell(\cdot,\cdot)$ is a strictly increasing monotonic hinge alignment function, where $\ell(a, b) = \max\big(0,\; m - \operatorname{sign}(a)\, b\big),
 m \ge 0.$

\noindent\textbf{Magnitudes}: more increase in \textit{progress} indicates more rewards.
Reward learning from human demonstrations relies more on distinguishing between policies covered by the demonstrations or policies outside the demonstrations, resulting in the learned reward values tending to be binary based on the similarity to expert behaviors.
However, 
even in optimal demonstrations, not all behaviors in a demonstration are equally critical.
Some steps, like picking up the target object, are inherently more progressive than simply moving closer, and thus should intuitively receive a larger reward.
So transitions with larger progress improvements should receive larger rewards. 
We enforce this by:
\begin{equation}
\label{eq:pair_align}
\begin{aligned}
\mathcal{L}_{\Delta p \rightarrow \Delta r_\theta}(\theta)
&=
\mathbb{E}_{(s_i,s_{i+1}, s_j, s_{j+1})\sim\mathcal{D}}
\Bigl[\ell\!\Bigl(
    \Delta p_i - \Delta p_j,\, \\
&\qquad\qquad
    r_\theta(s_i,s_{i+1}) - r_\theta(s_j,s_{j+1})
\Bigr)
\Bigr].
\end{aligned}
\end{equation}
Note that in this condition specifically, $s_i, s_{i+1}, s_j$, and $s_{j+1}$ are preferably from the same demonstration $D$ since different users might scale \textit{progress} differently.

\noindent\textbf{Potential: high \textit{progress} means high potential.}
The potential function $\Phi$ was introduced in \cite{ng1999policy}, and describes the distance between the current state and the goal state:
$\Phi(s) = -dist(s, s_{goal})$.
The use of potential function will not alter the original optimal policy, and can reduce random exploration by getting heuristics from: 
$F(s_t,a_t,s_{t+1})
\;=\;
\gamma\,\Phi(s_{t+1})
\;-\;
\Phi(s_{t})
$.
We set $\Phi(s) = V(s)$ in this work.
If a state has higher \textit{progress}, the learned potential should also be higher. We capture this by:
\begin{equation}
\label{eq:pot_align}
\begin{aligned}
\mathcal{L}_{\Delta p \rightarrow \Phi}(\Phi)
&=
\mathbb{E}_{(s_t,s_{t+1})\sim\mathcal{D}}
\Bigl[ \ell\bigl(\,\Delta p_t, F(s_t, a_t, s_{t+1}) \bigr)
\Bigr].
\end{aligned}
\end{equation}

\noindent\textbf{Completion: success represents more total rewards than failure.}
The ultimate goal of reward learning is to learn a reward function that guides the agent to complete the task, and thus, task completion imposes a total reward constraint. 
We therefore hypothesize that any demonstration $d_s \in \mathcal{D}_{\text{succ}}$ that completed the task should receive a higher total reward $
R_\theta(d) \;=\; \sum_{t} r_\theta\!\bigl(s_t\,s_{t+1}\bigr)$
 than any demonstration $d_{f} \in \mathcal{D}_{\text{fail}}$ that did not complete the task (in this work, $progress <$ 90 at the end) , no matter the horizon of the demonstrations.  
We capture this condition by:
\begin{equation}
\label{eq:rank_loss}
\begin{aligned}
\mathcal{L}_{\mathrm{comp}}(\theta)&=
\sum_{d_f\in\mathcal{D}_{\mathrm{fail}}}
\max\Bigl\{0,\,R_\theta(d_f) \\
&\qquad - \frac{1}{|\mathcal{D}_{\text{succ}}|} \sum_{d_s \in \mathcal{D}_{\text{succ}}} R_\theta(d_s)\Bigr\},
\end{aligned}
\end{equation}
where $\mathcal{D}_{\text{succ}}$ are successful demonstrations  and $\mathcal{D}_{\text{fail}}$ are unsuccessful demonstrations.

\subsection{PHIRL}
In PHIRL, we learn reward functions by alternately updating a reward function by using AIRL and aligning the learned reward function with \textit{progress}.
Given demonstration dataset $\mathcal{D}$ and their annotated subset $\mathcal{D}_p$, a learned reward $r_\theta$ is \emph{aligned} by minimizing a sum of all shaping components:         
\begin{equation}
\begin{aligned}
\mathcal L_{\mathrm{align}}(\theta, \Phi)
&=
\lambda_1
\mathcal L_{\Delta p \rightarrow r_\theta}(\theta)
+
\lambda_2
\mathcal L_{\Delta p \rightarrow \Delta r_\theta}(\theta) \\
&\quad +
\lambda_3
\mathcal L_{\Delta p \rightarrow \Phi}(\Phi)
+
\lambda_4
\mathcal L_{\mathrm{comp}}(\theta),
\end{aligned}
\label{eq:weighted_alignment_objective}
\end{equation}      
where we configured $\lambda_1 = \lambda_2 = \lambda_3 = 1$ and $\lambda_4 = \frac{1}{|\mathcal{D}|}$. 
In offline PHIRL, all demonstrations and annotations are collected before the learning phase.
We show PHIRL in Algorithm \ref{alg:PHIRL}.

\begin{algorithm}[htbp]
\caption{PHIRL}
\label{alg:PHIRL}
\begin{algorithmic}[1]
    \State  Collect demonstration dataset $\mathcal{D}$, annotate a subset $\mathcal{D}_p$, initialize discriminator $D_{\theta}$ and policy $\pi$
    \While{not converged}
        \State Sample trajectories $\tau \sim \mathcal{D}$ and $\tau_\pi \sim \pi$, update $D_{\theta}$ to distinguish $\tau$ from $\tau_\pi$
        \State Sample labeled trajectories $\tau_p \sim \mathcal{D}_p$ and corresponding labels $p_{\tau}$
        \State Compute reward $r_{\theta}(\tau_p) $, potential $\Phi(\tau_p)$
        \State Update $r_{\theta}$ and $\Phi$ with respect to updated $p$ by minimizing $L_{align, \theta, \Phi}$
        \State \textbf{Optional} annotate $\tau_\pi$ with VLM-generated \textit{progress}, update $\mathcal{D}_p \leftarrow \mathcal{D}_p \cup \{\tau_\pi\}$
    \EndWhile

\end{algorithmic}
\end{algorithm}

\noindent \textbf{Online Extension} While human-in-the-loop teaching is \textit{not mandatory} for our method, PHIRL can benefit from adding generated demonstrations into the \textit{annotation dataset}. The only change to the method for online learning is that $\mathcal{D}_p$ is updated throughout learning with new annotated demonstrations chosen from generated trajectories.  PHIRL Online can improve the robustness of the learning outcome, and better align the reward function with actual human intentions. 

\noindent \textbf{Scalability} We choose to apply PHIRL to AIRL over other IRL algorithms because AIRL's training objectives are well-aligned with \textit{progress} in multiple aspects, but the PHIRL framework itself is not inherently tied to AIRL. For instance, for methods without a potential component, PHIRL can still be effective with the remaining three alignments (see \autoref{sec:abl}). The key idea of the framework is shaping the learned rewards to align with the evaluation information specified by \textit{progress}.

\section{Experiments and Results}
\label{sim exp}
In this section, we run PHIRL with four baselines using one task with a real robot and two tasks from Robosuite \cite{zhu2025robosuitemodularsimulationframework} to validate PHIRL's performance and our claimed contributions.
In addition, we demonstrated that the reward functions learned by PHIRL are robust to reward hacking in two representative scenarios.

\subsection{Setups}
\noindent \textbf{Domains}
We selected two domains from Robosuite, Lift and PickPlaceCan, and one real robot task, animal-brushing.
For the Lift domain, the goal is to reach the block and lift it to a certain height. 
For the PickPlaceCan, the goal is to pick up the can, navigate to the goal location, and drop the can into the right bin.
For the real robot task, we used a Kinova Gen 3 Lite arm and an animal-brushing brush task. The goal was to navigate to the stuffed animal and then brush the pom-poms off the stuffed animal.

\noindent \textbf{Datasets}
For the Robomimic domains, we trained baselines and PHIRL on a high-quality dataset and a mixed-quality dataset.
The high-quality datasets are the Proficient Human (PH) datasets, which were provided by a single proficient operator. Each dataset consists of 200 demonstrations. 
The mixed-quality datasets are the Multi Human (MH) datasets, which were collected by six operators using the RoboTurk platform. 
Each operator provided 50 demonstrations. 
Two operators were ``worse'' operators, two were ``okay'' operators, and two were ``better'' (i.e., proficient) operators.
For the real robot domain, we used a similar setup, recruited six participants (two ``worse'', two ``okay'', and two ``better'', 10 demonstrations each), and constructed an MH dataset.

\begin{figure*}[!t]
    \centering
    \includegraphics[width=1\linewidth]{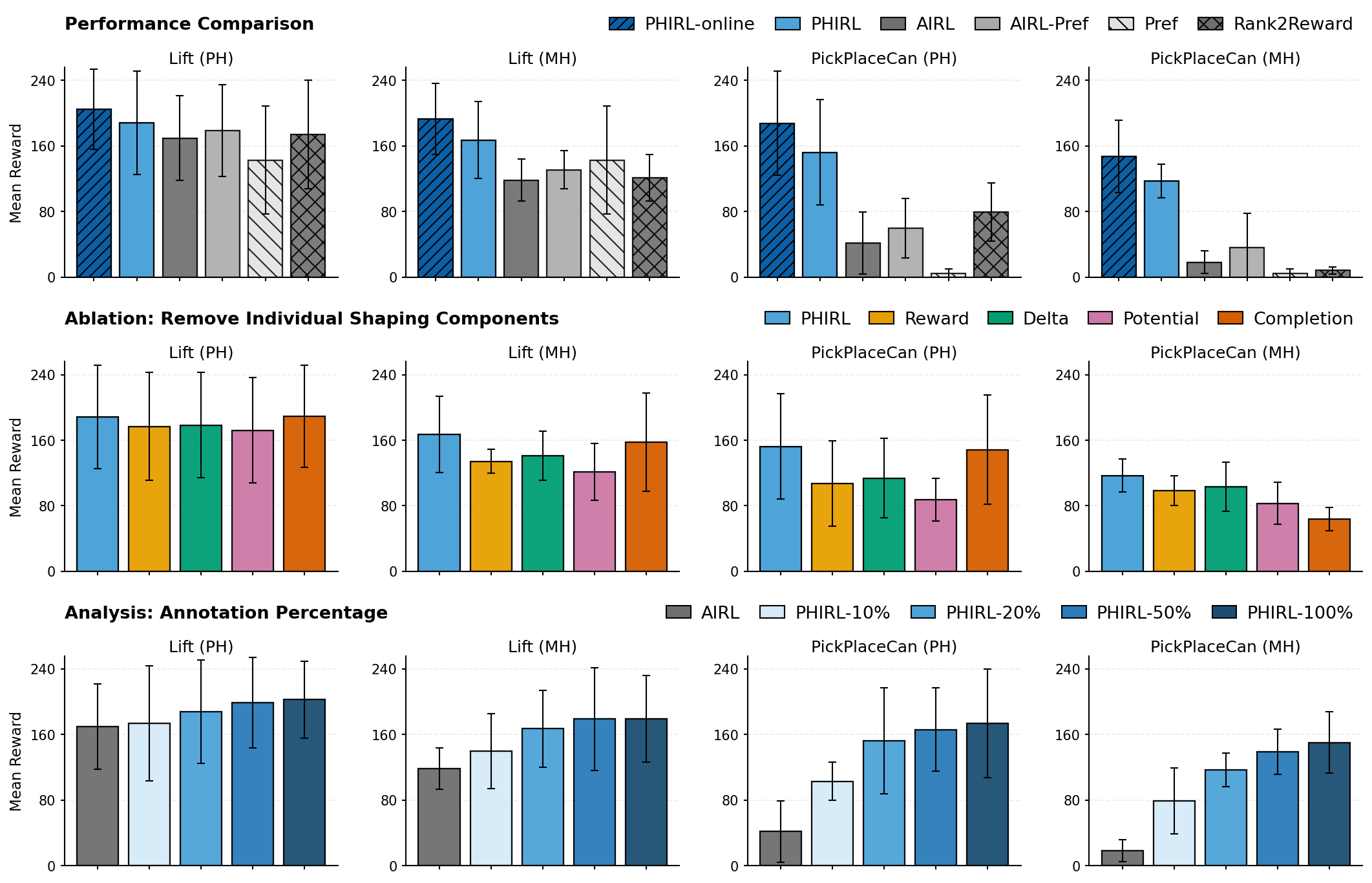}
    \caption{\textbf{Performance comparison and ablation studies} 
    Results in the Lift and PickPlaceCan domains.
    Error bars represent standard deviation.
    Note that the standard PHIRL model uses 20\% annotation (equivalent to PHIRL-20\%). 
    \textbf{(Top)} Average environmental return rewards across all baselines. 
    PHIRL significantly outperforms baselines in both domains, particularly when the task horizon is long or demonstration quality is low (MH). 
    \textbf{(Middle)} Ablation study on the effect of removing individual reward shaping components. 
    Completion shaping is most critical when there are failed demonstrations, while reward shaping is most effective overall. 
    \textbf{(Bottom)} Sensitivity analysis of annotation percentages. PHIRL improves learning significantly with 20\% of the demonstrations. 
    The benefits of the offline annotation saturate at 50\% if the demonstrations are relatively high quality.  
    }
    \label{fig:res}
\end{figure*}

\noindent
\textbf{Baselines} 
We include four baseline methods, AIRL (IRL with no human feedback) \cite{fu2018learning}, Pref (preference-learning based IRL with no demonstrations) \cite{christiano2017deep}, AIRLPref (IRL with both feedback and demonstrations) \cite{palan2019learning}, and Rank2Reward (reward learning method with a progress-ish signal) \cite{yang2024rank2reward}. 
We use PPO \cite{schulman2017proximalpolicyoptimizationalgorithms} as the adversarial learners and trajectory generators. 
We set the standard PHIRL model to use a 20\% annotated demonstration dataset. 
We trained all baselines and PHIRL for 500 episodes, with 10 parallel environments and an environmental horizon of 256. 
Hyperparameters, including learning rate, discount factor, clip range, and discriminator batch size, are selected by using a hyperparameter search \cite{optuna_2019}.

\noindent \textbf{Data Collection} We separately recruited six participants to annotate the Robomimic datasets, six participants to construct the real robot dataset, and two participants to annotate the real robot dataset. 
When collecting \textit{progress} and preferences, each demonstration was equally divided into 10 sub-trajectories. Each sub-trajectory was annotated with one \textit{progress} label, and for preferences, we asked participants to pick a preferable sub-trajectory from two sub-trajectories.
To reduce bias in \textit{progress} annotation, we avoided imposing specific patterns on the \textit{progress} values by only asking them to provide values from 0 to 100. 
The VLM base model used in this work is Gemma 3 12B \cite{Gemma3Team2025}, we fine-tuned the base model using LoRA \cite{hu2021loralowrankadaptationlarge}.  
More details about the implementation,
 prompts used, and the code can be found at  github.com/PHIRL2026/PHIRL\_appendix.

\begin{figure}[!t]
    \centering
    \includegraphics[width=\linewidth]{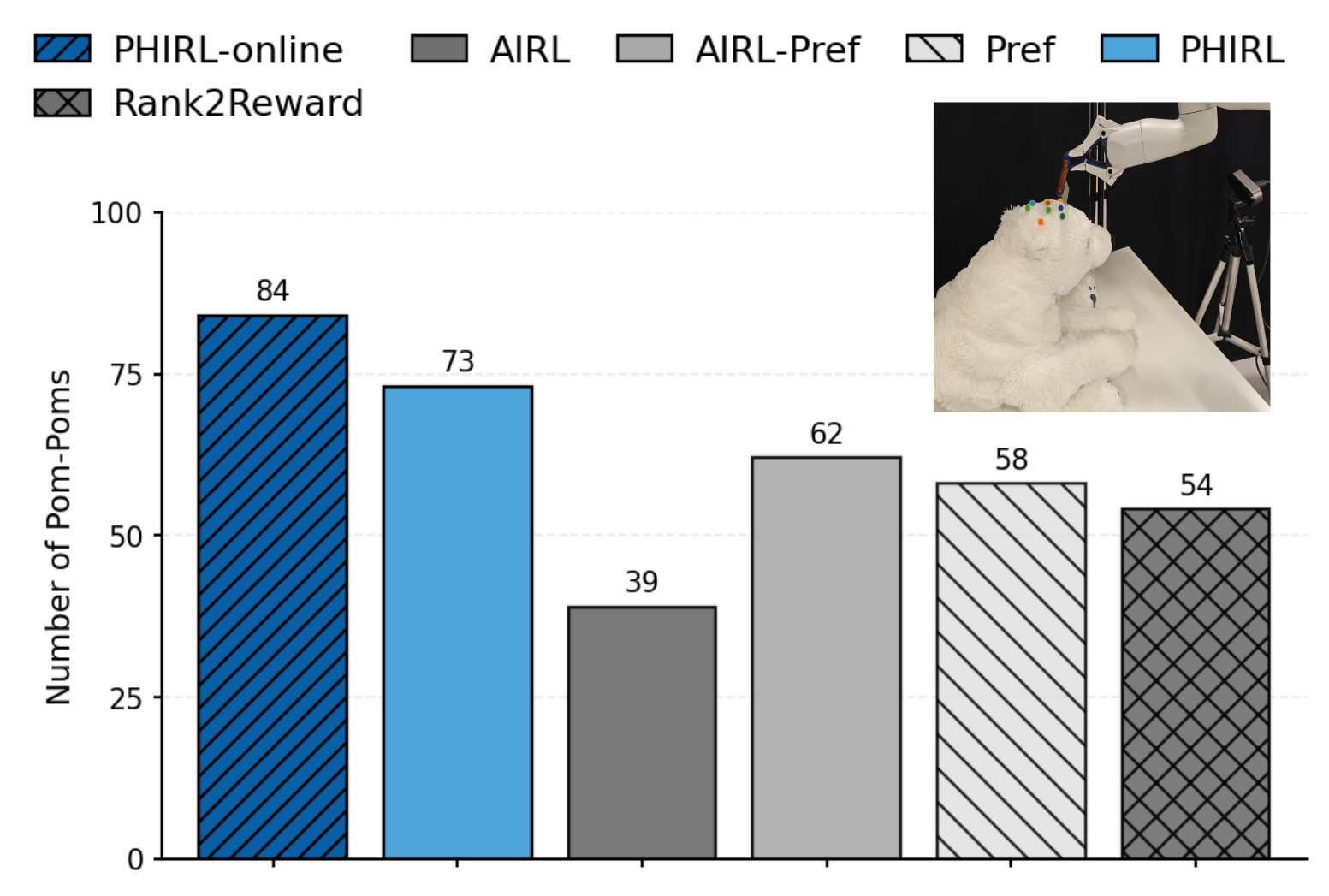}
    \caption{Real Robot Study Results. The goal is to brush the pom-pom off the stuffed animal. We calculated the number of pom-poms that were brushed off the stuffed animal. PHIRL-online achieves the highest.}
    \label{fig:rrs}
\end{figure}

\begin{figure*}[!t]
\label{fig:lift_sce}
    \centering
    \includegraphics[width=1\linewidth]{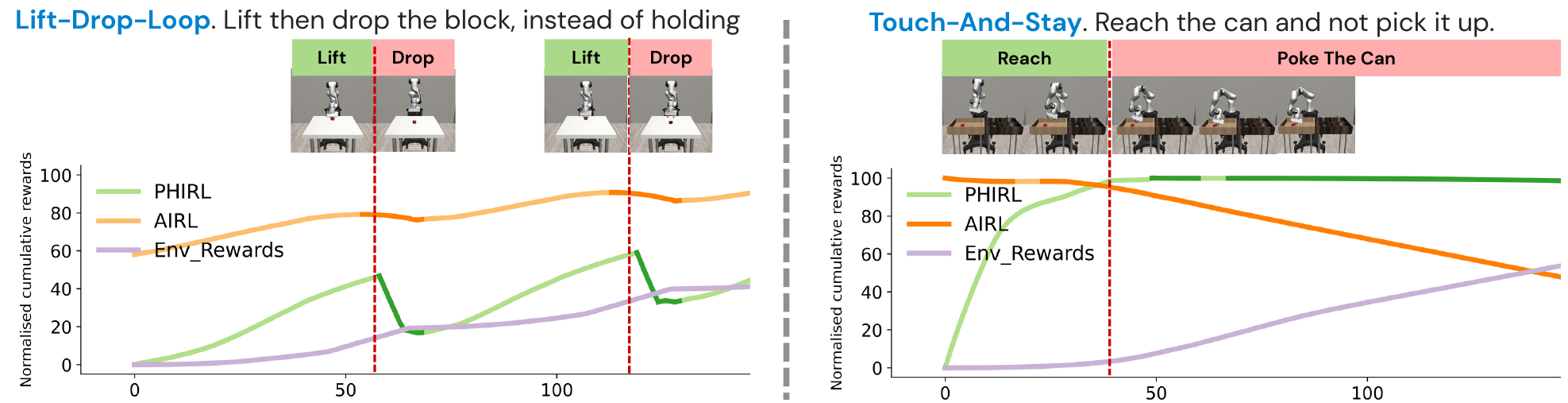}
    \caption{
    \textbf{Reward Hacking Scenarios and Reward Curves.}
    The lighter colors and darker colors indicate increases and decreases in rewards.
    The Lift-Drop Loop  was generated by a PPO agent that learns from AIRL rewards. The robot repeatedly lifted the block and then dropped it.
    We found that AIRL encouraged the robot to collect more rewards by lifting and dropping the block repeatedly,
    while PHIRL appropriately punished dropping behaviors and rewarded lifting behaviors. 
    The Touch-And-Stay was generated by a PPO agent that learns from Robosuite environmental rewards. The robot navigated to the can and kept poking it instead of picking and placing it.  PHIRL gave a negative but near-zero reward, indicating the robot is not making progress. }
    \label{fig:rh_sce}
\end{figure*}

\subsection{PHIRL Outperforms Baselines}
\textbf{Simulation.}
We trained baselines and PHIRL on a high-quality dataset (PH) and a mixed-quality (MH) dataset from Robomimic \cite{mandlekar2021matterslearningofflinehuman}. 
The results of simulation experiments are presented in Figure \ref{fig:res}. 
The top row presents the average environmental return rewards.
Since Pref does not learn from demonstrations, we used the same data for both PH and MH conditions.
Error bars and tests are computed over 100 evaluation rollouts (environment seeds 1 to 100).
We used Welch’s t-test to determine statistical significance.
In the Lift domain, PHIRL achieved a mean environmental reward of $M=188.12$ in PH and $M=167.16$ in MH, which is higher than all other baselines (5.6\% and 17.1\% higher than the best baseline). 
In the more challenging PickPlaceCan domain, where the average demonstration horizon is approximately twice that of Lift, PHIRL achieved greater performance advantages (91.3\% higher than the best baseline) and also became more significant in the MH dataset(2.3 times more return rewards), where demonstration quality is lower. 
These results demonstrate that PHIRL learns effective rewards for RL agents, and is more reliable when the task horizon is longer or the demonstration quality is lower.

\textbf{Real Robot} 
For the real robot domain, we recruited six participants and constructed an MH dataset in a similar way (two ``worse'', two ``okay'', and two ``better'', 10 demonstrations each). 
We trained PHIRL and baselines over the constructed dataset and calculated the number of pom-poms being brushed off from the stuffed animal over 10 episodes. 
For each episode, 10 pom-poms were placed randomly. 
Results are shown in \autoref{fig:rrs}.
Our results confirmed our findings in simulations. PHIRL performed well with mixture quality demonstrations and outperformed baselines AIRL (39), AIRL-Pref (62), Pref (58), and Rank2Reward (54).

\noindent \textbf{PHIRL vs. PHIRL-online}
We found that using VLM-generated \textit{progress} labels (400 for PH and 600 for MH in addition, the same number as human annotations) and for online training resulted in a consistent improvement for PHIRL. 
In the Lift-PH condition, we found that adding online progress labels resulted in a statistically significant improvement at the converged performance between PHIRL and PHIRL Online ($M=188.12$ vs $M=205.08$, $p < 0.05$).
PHIRL-online also demonstrated a decent improvement on the real robot study (84 vs. 73). 
The impact of online labels became even more pronounced when the demonstrations were noisier (MH settings) or the task was harder (PickPlaceCan Domain). For instance, in the challenging Can-MH scenario, returns increased from $M=116.99$ to $M=147.28$ ( $p < 0.001$).

\subsection{Ablation Study and Sensitivity Analysis} 
\label{sec:abl}
\noindent{\textbf{Ablation.}} We investigated the contribution of each individual reward-shaping term to the overall reward function, as shown in the middle row of \autoref{fig:res}. We treat the standard PHIRL model as the anchor. 
The potential shaping component is generally influential across all conditions.  
Other ablations also showed statistically significant decreases, indicating the full suite of reward shaping components is necessary for maximal performance.
The only exception is the completion shaping term, as  in most conditions demonstrations are marked as fully completed (\textit{progress} $>$ 90 for the last frame).

\noindent{\textbf{Sensitivity.}} The bottom row of \autoref{fig:res} shows the sensitivity of PHIRL to the annotation coverage. Our standard PHIRL utilizes 20\% of the available annotations. 
The results indicate that increasing the coverage in general will increase the model performance. 
The improvement from annotation is more significant when the dataset has a lower quality.
In most conditions, the improvements saturated at 50\% of coverage, suggesting the strong data efficiency of PHIRL.

More details and results, including more statistical analysis, learning curves, success rate, annotation examples, and additional domains for simulation studies, are in github.com/PHIRL2026/PHIRL\_appendix.

\subsection{PHIRL Learns More Robust Reward Functions}
When training PHIRL, AIRL, and PPO, we found two reward hacking scenarios: Lift-Drop-Loop and Touch-And-Stay.
We show the keyframes of the reward-hacking trajectories and cumulative rewards for PHIRL, AIRL, and PPO in the Robosuite environment. 
The rewards are normalized to 0 to 100 for visualizing the curves. 
We use a lighter color to indicate increases in rewards and a darker color to represent decreases in rewards.

\noindent \textbf{Lift-Drop-Loop}
We show keyframes of the trajectory and a normalized reward curve for each method in \autoref{fig:rh_sce}. 
This scenario was generated by a PPO agent that learns from the AIRL reward function. 
The robot repeatedly picks up the block, lifts it, and then drops it. 
This is because AIRL rewards the robot for picking up the block, but does not continue to reward or even punish the robot for having the block once it has been picked up. Thus the robot can achieve a higher cumulative reward by repeatedly lifting and dropping. 
PHIRL produced a more robust reward function. It positively rewarded the repeated reaching and lifting behaviors, and punished the dropping behaviors since dropping the block back to the table brings the state back to a low \textit{progress} status.

\noindent \textbf{Touch-And-Stay}
The scenario was produced by a PPO agent that learns from the environmental rewards, shown in \autoref{fig:rh_sce}. 
Instead of picking up the can and placing it in the target bin, the robot navigated to the can, closed the gripper, and poked the can with the gripper. 
This reward hacking is likely caused by an imbalanced reward design and a bottleneck in the task. 
Grasping the can is challenging and has risks of losing future rewards, so the PPO agent learned a ``safer'' strategy, collecting small rewards by poking the can.
Without carefully tuning the hyperparameters, the agent can easily get stuck in this local optimum. 
In contrast, AIRL and PHIRL both negatively rewarded the poking behaviors, while PHIRL gave a very small negative reward since poking behaviors are not harmful to the task completion.

\section{Discussion and Limitation}

 \textbf{Sampling Annotated Demonstrations}
In this work, when we trained PHIRL with partially annotated datasets, we randomly sampled demonstrations from the dataset instead of intentionally selecting imperfect demonstrations.  We believe that the process of classifying human demonstrations into high-quality and low-quality introduces additional information to the learning process, which can bias our comparison with baselines. 
However, we agree that preferentially selecting lower-quality demonstrations will improve sample efficiency and thus improve model performance, especially when there are limited human feedback labels.

\begin{figure}[htbp]
    \centering
    \includegraphics[width=0.99\linewidth]{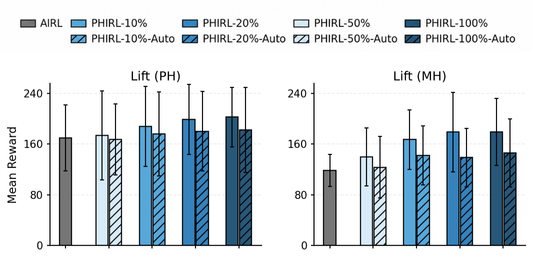}
    \caption{ Comparisons Between Human Annotations and Auto-Generated Annotations.  We compare the performance of the model at different annotation coverages.
    The auto-generated annotations from an unfine-tuned VLM can improve learning from demonstrations with PHIRL.
}
    \label{fig:navie}
\end{figure}

\textbf{Exploring Generating Progress Labels without Human Annotations with A VLM} 
We conducted a formative study to explore generating progress labels without any human annotations.
We used a ReWiND-style method \cite{zhang2025rewind} and a Gemma 3 12B model. 
We took the key idea of ReWiND: “use language instructions to represent key subgoals in a task, and then combine the language instructions with a VLM to indicate task progress.” We modified the prompt we used for the fine-tuned model, specify the subgoals with instructions, and use an unfine-tuned Gemma 3 12B to annotate the human demonstrations. 
The results are in \autoref{fig:navie}.

\vspace{-10pt}

Our results show that generated annotations from an unfine-tuned VLM can improve learning with PHIRL in most cases (ph: 169.62 vs. 167.33, 175.80, 180.13, and 182.54, and mh: 118.30 vs. 123.35, 141.92, 138.52, and 145.91), despite the performance is lower than using real human feedback.
While the performance gain is less significant and saturates faster than with real human feedback,
it demonstrates that PHIRL and other progress-based learning algorithms can benefit from a VLM trained with general knowledge.

\textbf{Limitations and Future Work} 
We acknowledge several limitations in our work:
1. Our method assumes that the task has a finite horizon or a complete status, making it unsuitable for tasks with an infinite horizon or cyclic observations, such as ant\_walk from DeepMind Control Suite \cite{tassa2018deepmind}. 
2. Methods are evaluated in one training run, as AIRL-based methods are generative adversarial inverse reinforcement learning methods that learn online with adversarial training; repeating every condition with multiple training seeds was beyond our budget. 
3. As we discussed earlier, we randomly sampled demonstrations from the dataset for annotations rather than actively selecting imperfect demonstrations.
We expect that preferentially selecting lower-quality demonstrations will improve the sample efficiency, and thus improve model performance, especially when the number of human feedback labels is limited. 
PHIRL currently relies on \textit{progress} feedback. Recent advances show that semantic instructions from humans \cite{gunjal2025rubrics,liu2025openrubrics} or VLMs \cite{ma2024visionlanguagemodelsincontext, bu2025univla, shukla2023lgtsdynamictasksampling} can serve as compelling modalities for reward generation and alignment. One future work is to improve the learned function by blending trajectory-level progress with language instructions and structured rubrics.

\section{Conclusions}
In conclusion, we proposed PHIRL, an inverse reinforcement learning method that learns a reward function by alternately inferring a reward function and aligning the inferred reward function with \textit{progress} annotations over annotated demonstrations. 
PHIRL is flexible to varying human participation levels since the annotations can be collected before or during learning.
We showed that PHIRL significantly outperforms baselines with 20\% of demonstrations being annotated before learning.
We also showed that VLM-generated \textit{progress} could enhance PHIRL's converged performance. 
PHIRL exhibited a robust performance in two reward hacking scenarios, Lift-Drop-Loop and Touch-And-Stay. 
We believe that PHIRL  is an effective framework for learning robust reward functions, and the intuition of aligning the learned policy with human annotation is an efficient way of leveraging human demonstrations and human feedback.

\bibliographystyle{IEEEtranN}   %
\bibliography{example}  %

\end{document}